%% file: 00_main.tex
\documentclass[runningheads]{llncs}
\usepackage[T1]{fontenc}
\usepackage{graphicx}
\usepackage{xcolor}
\usepackage{multirow}
\usepackage{ulem}

\usepackage{xspace}

\newcommand{\framework}{\textit{CogApp}\xspace}
\newcommand{\arch}{\textit{CogGen}\xspace}

\begin{document}
%
\title{CogGen: A Learner-Centered Generative AI Architecture for Intelligent Tutoring with Programming Videos}

%
\titlerunning{CogGen: Intelligent Tutoring with Programming Videos}

\author{Wengxi Li\inst{1}\orcidID{0009-0003-1876-6303} \and
Roy Pea\inst{2}\orcidID{0000-0001-6301-3536} \and
Nick Haber\inst{2}\orcidID{0000-0001-8804-7804} \and
Hariharan Subramonyam \inst{2}\orcidID{0000-0002-3450-0447}}
\authorrunning{Wengxi Li et al.}
%
\institute{City University of Hong Kong, Hong Kong SAR, China\\
\email{wengxili@cityu.edu.hk}\\
\and
Stanford University, Stanford, California, United States\\
\email{\{roypea,nhaber,harihars\}@stanford.edu}}
\maketitle              
\begin{abstract}

We introduce CogGen, a learner-centered AI architecture that transforms programming videos into interactive, adaptive learning experiences by integrating student modeling with generative AI tutoring based on the Cognitive Apprenticeship framework. The architecture consists of three components: (1) video segmentation by learning goals, (2) a conversational tutoring engine applying Cognitive Apprenticeship strategies, and (3) a student model using Bayesian Knowledge Tracing to adapt instruction. Our technical evaluation demonstrates effective video segmentation accuracy and strong pedagogical alignment across knowledge, method, action, and interaction layers. Ablation studies confirm the necessity of each component in generating effective guidance. This work advances AI-powered tutoring by bridging structured student modeling with interactive AI conversations, offering a scalable approach to enhancing video-based programming education.

\keywords{Cognitive Apprenticeship  \and Large Langauge Models \and Student Modeling \and Conversational Agents}
\end{abstract}
\input{01_intro_new}
\input{03_methods_new}
\input{04_results_new}
\input{05_discussion_new}

%
%
%
%
\bibliographystyle{splncs04}
\bibliography{99_refs}




\end{document}

%% file: 01_intro_new.tex
\section{Introduction}
Video has emerged as a preferred medium for learning programming. According to a recent Stack Overflow survey, developers who learned through ``How-to-videos'' and ``Video-based E-Courses'' accounted for 54.2\% and 49.9\%, respectively~\footnote{https://survey.stackoverflow.co/2024/developer-profile\#learning-to-code}. Regardless, the most effective learning paradigm for programming videos is \textit{learning by doing}~\cite{Brown1989SituatedCA}. However, practicing coding tasks while watching videos is difficult for novices, requiring clear instructions and immediate feedback~\cite{alario2016interactive}.

Learning to program requires integrating multiple skills: understanding syntax, grasping programming concepts, and applying problem-solving strategies~\cite{linn1995designing}. \textit{Instructional Design Principles} guide how these learning experiences should be structured~\cite{gagne2005principles}. The Cognitive Apprenticeship (\framework) framework addresses these needs by making expert programmers' thought processes visible to novices through six teaching moves: Modeling, Coaching, Scaffolding, Articulation, Reflection, and Exploration~\cite{collins1991cognitive}. This framework emphasizes guided practice with gradual transfer of responsibility to learners, effectively managing cognitive load through step-by-step instruction~\cite{sweller2021instructional,anderson2001taxonomy,Pea2004TheSA}.

While large language models (LLMs) have shown promise in programming education~\cite{liu2024teaching,10.1007/978-3-031-64302-6_19},  they present several challenges: they tend toward verbose responses~\cite{zhao2023survey,cheng2024scientific}, struggle to maintain focus in multi-turn conversations~\cite{sreedhar-etal-2024-canttalkaboutthis,sheng-etal-2023-dialogue}, and lack structured approaches for step-by-step teaching~\cite{ren2024human,sonkar-etal-2023-class}.




To address these gaps, we developed \arch, a systematic architecture that ensures controlled and relevant message generation tailored to students' specific learning needs based on the \framework framework. \arch operates by first segmenting programming videos into distinct learning goals, then selecting appropriate teaching methods from the \arch framework based on an internal student model for each learning goal. In summary, our key contributions include:



%
\begin{enumerate}
 \item \arch: A pipeline for generating learning conversations, using a domain-specific language to organize \framework methods, enabling LLMs to deliver guidance that is both consistent and personalized.
 \item Technical evaluation demonstrating \arch's effectiveness in generating structured and controlled instruction via LLMs, with evidence that each component plays a necessary role in achieving this performance.
\end{enumerate}

%% file: 03_methods_new.tex
\section{\arch Architecture}

\arch implements a set of LLM prompting techniques that yield conversational utterances grounded in the \framework framework. The architecture consists of (1) a video segmentation module that \textit{slices} long videos based on learning goals using a prompt chaining approach, (2) a DSL generator that produces the \textit{context} for the LLM to generate utterances in interaction with the learner, and (3) a student model to capture the student's learning progress and inform appropriate mentor \textit{moves}. Here, we provide technical details about each module and data representations to make the generated conversations \textit{consistent} and \textit{controllable}. We use few-shot prompting with GPT-4 (temperature=0.3) for all the modules.



\subsection{Video Segmentation by Learning Goals}

When segmenting video transcripts, small references to unrelated goals can cause misclassification. Our approach minimizes this through a three-step process: (1) \textit{Summarize} key points for each learning goal; (2) \textit{Retrieve} transcript sentences aligning with these summaries; (3) \textit{Rearrange} segments based on timestamps.

\subsection{Instructional Prompt Generator}
For each video segment, we generate a teaching plan that aligns with the \framework framework. Our prompting pipeline follows three consecutive steps:

\begin{table*}[t!]
\centering
\renewcommand{\arraystretch}{1.5}
\caption{Representations of the two types of knowledge in \arch. The bold parts are later used in the student model.}
\begin{tabular}{p{2cm}|p{4.5cm}p{5.5cm}}
\hline
\textbf{Knowledge} & \textbf{Representation} & \textbf{Example} \\ \hline
\multicolumn{3}{c}{\textit{Concept Related}} \\ \hline
Declarative & [Subject] + [verb phrase] + that + [independent clause]. & \textit{The median income by college major shows that majors earn a median income of over \$30K right out of college.} \\ \hline
Procedural & To achieve/understand + [specific goal/outcome] + one must + \textbf{[actions/processes]} + [additional details] + considering/using + [relevant factors/tools]. & \textit{To understand the distribution of earnings by college major, one must \textbf{examine the histogram and identify overall trend or extreme values}, considering whether high earnings are due to the field's financial reward.} \\ \hline
\multicolumn{3}{c}{\textit{Programming Related}} \\ \hline
Declarative & The task is + [final goal] + using + [general method/tool] + and + [additional method/technique for enhancement]. & \textit{The task is comparing the distribution of median earnings across different major categories using a box plot and adjusting the visualization for better readability and interpretation.} \\ \hline
Procedural & To achieve + \textbf{[specific goal]} + one must + \textbf{[action/verb] + [specific tool/method]} + on + [object/target] + because + [reason/purpose]. & \textit{To \textbf{achieve an ordered factor level} based on the `Median', one must \textbf{use `fct\_reorder'} on `Major\_category', making it easier to compare distributions.} \\ \hline
\end{tabular}
\label{tab:knowledge}
\end{table*}

\begin{figure*}[t!]
  \centering
  \includegraphics[width= \textwidth]{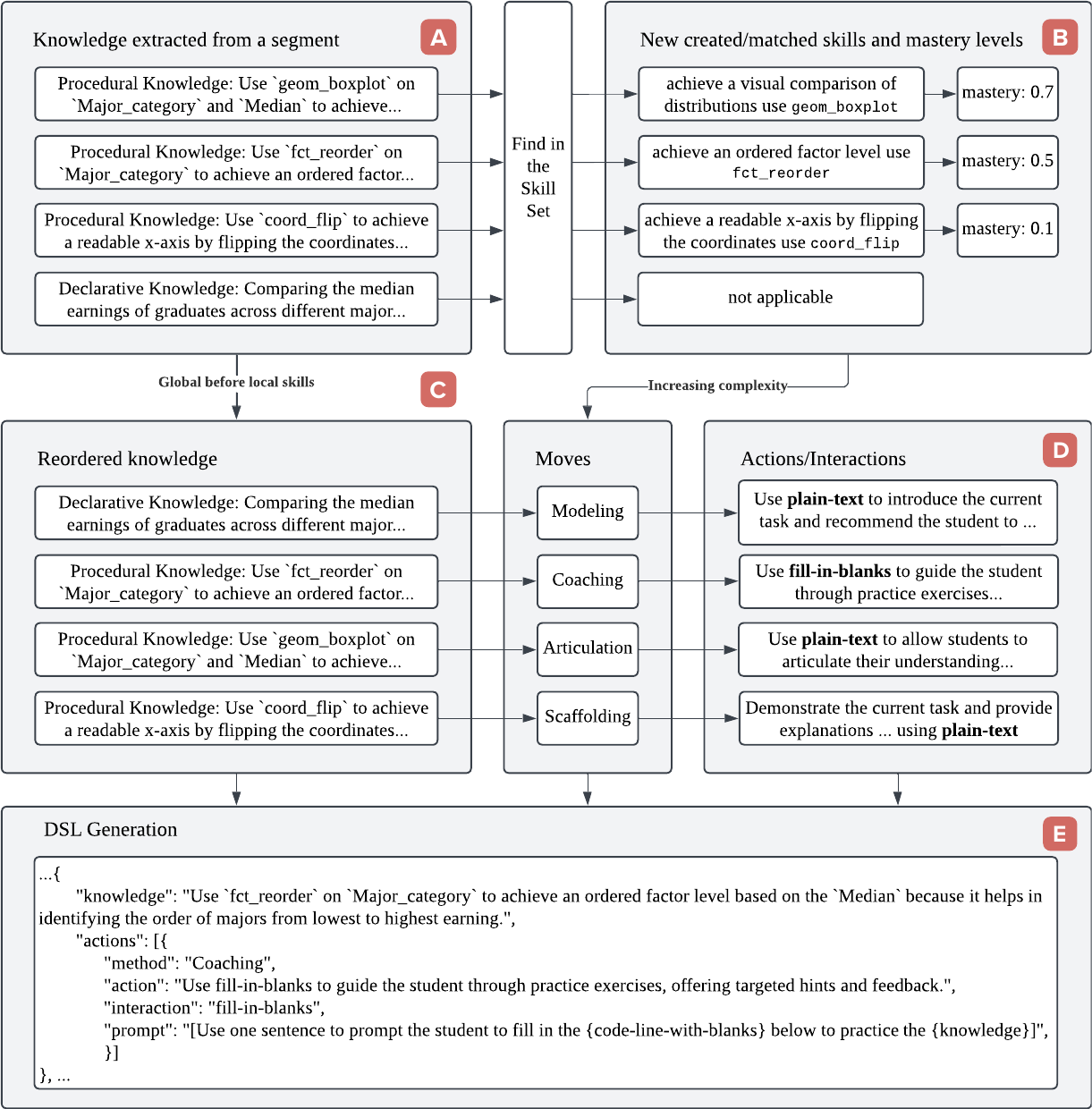}
  \caption{Details of the \arch architecture: (A) extracting procedural knowledge and declarative knowledge from the video segment, (B) the knowledge is mapped to skills and mastery levels, (C) using the principles to reorder the knowledge and decide moves for each knowledge, (D) moves are mapped to actions and interactions, (E) knowledge, moves, actions, interactions, and prompts are integrated to form the DSL.}
  \label{fig:detail}
\end{figure*}

First, we use few-shot prompts to extract procedural and declarative knowledge from the segmented transcripts using standardized formats (Figure~\ref{fig:detail}A and Table~\ref{tab:knowledge}). For example, procedural knowledge follows patterns like ``To achieve [specific goal] one must [action/verb] [specific tool/method] on [object/target] because [reason/purpose].'' This structured approach enables mapping to the student model and ensures consistent generation of conversational messages.


Second, we apply three principles to select proper teaching methods~\cite{collins1991cognitive}: (1) global before local skills, prioritizing conceptual overviews before detailed tasks; (2) increasing complexity, adapting instruction to student mastery levels; and (3) increasing diversity, varying task types to maintain engagement.(Figure~\ref{fig:detail}B\&C)

Finally, we generate a domain-specific language (DSL) that integrates knowledge summaries, pedagogical moves, and context to produce structured interactions (Figure~\ref{fig:detail}D\&E). Each move (e.g., Modeling, Scaffolding) is mapped to specific actions and interaction types, which are used to generate \textit{prompts} that guide the conversation. The prompts are stored in a queue, the head of which is used to produce the next message. This approach ensures interactions remain focused on the video content while following the intended pedagogical sequence.

\subsection{Student Performance Monitoring} \label{sec:studentmodel}

To adapt instruction based on student performance, we implement a Bayesian Knowledge Tracing (BKT) framework that tracks skill proficiency~\cite{badrinath2021pybkt,10.1145/3544548.3581574}. Each knowledge component extracted from video segments is translated into a corresponding student skill (e.g., ``achieve an ordered factor level use \texttt{fct\_reorder}'').

Skill proficiency levels are initialized based on the student's estimated mastery of the current learning goal (default 0.1). As students interact with the system, BKT parameters are dynamically updated to reflect evolving proficiency. When students encounter previously practiced knowledge, we use semantic similarity~\cite{youdao_bcembedding_2023} to identify relevant skills and update their parameters. The parameters are stored between learning sessions to maintain continuity across videos.

This integrated approach allows CogGen to select appropriate teaching methods based on current student mastery, providing scaffolding for low-mastery skills and encouraging articulation for higher-mastery skills, creating a personalized learning experience that adapts to individual student needs.

%% file: 04_results_new.tex
\section{Evaluation}

To evaluate \arch's effectiveness across diverse learning scenarios, we selected three distinct programming video lesson topics: Exploratory Data Analysis (EDA), Machine Learning (ML), and Game Development. We evaluated the architecture using three metrics: (1) segmentation accuracy, (2) controllability of generated content, and (3) component importance through ablation studies.

\subsection{Procedure}
Two experts with experience in educational video analysis served as annotators for segmenting videos according to learning goals. For segmentation evaluation, we compared timestamps between manually labeled segments and those generated by LLM, using a five-second error threshold to account for differences between continuous video viewing and discrete transcript processing.

For the controllability evaluation, we created a hierarchical classification scheme based on the prompt pipeline. The system-generated data were extracted from DSL files, which defined expert-specified teaching methods. Three domain experts independently labeled 277 dialogue messages in six videos (124 EDA, 89 ML, and 64 Game Development utterances). We compared labeled data with system-generated data using precision, recall, and F1-score metrics.

To evaluate the component importance of \arch, we compared four conditions: (1) \textit{Baseline}, unconstrained generation using only video segments; (2) \textit{Knowledge-Only}, which summarizes segment knowledge without \framework theory; (3) \textit{Method-Only}, which uses \framework theory without specific knowledge awareness; and (4) \textit{Full} condition with the complete pipeline. Expert evaluators ranked the results in the dimensions of credibility, validity, and interactivity. The agreement between the evaluators was moderate (Spearman $\rho = 0.71$), and we calculated a TrueSkill score~\cite{herbrich2007trueskill} for each ablation, performed Kruskal-Wallis tests to determine the general differences between ablations, and applied Dunn's post hoc test~\cite{upton2014dictionary} to isolate which specific condition differed from another.

\subsection{Results}
\subsubsection{Segmentation Performance:} Video transcript segmentation achieved 76.9\% accuracy within the five-second threshold, providing acceptable performance for educational content. The accuracy decreased with longer videos, suggesting pre-segmentation into shorter clips (10-12 minutes) could improve results.
\begin{table*}[t!]
\centering
\caption{Comparison of the intent performance across three video topics: exploratory data analysis (EDA), machine learning (ML), and game development (Game).}
\begin{tabular}{|l|ccc|ccc|ccc|ccc|}
\hline
\multirow{2}{*}{Topic} & \multicolumn{3}{c|}{Knowledge} & \multicolumn{3}{c|}{Method} & \multicolumn{3}{c|}{Action} & \multicolumn{3}{c|}{Interaction} \\ \cline{2-13} 
                       & Prec.   & Recall  & F1   & Prec.  & Recall & F1   & Prec.   & Recall  & F1   & Prec.    & Recall   & F1    \\ \hline
Total                  & 0.791       & 0.787   & 0.789        & 0.814      & 0.809  & 0.807        & 0.902       & 0.895   & 0.896        & 0.970        & 0.968    & 0.968         \\
EDA                    & 0.810       & 0.806   & 0.808        & 0.849      & 0.815  & 0.818        & 0.896       & 0.871   & 0.872        & 0.984        & 0.984    & 0.984         \\
ML                     & 0.795       & 0.798   & 0.796        & 0.825      & 0.809  & 0.807        & 0.900       & 0.899   & 0.899        & 0.956        & 0.944    & 0.947         \\
Game                   & 0.827       & 0.781   & 0.792        & 0.808      & 0.813  & 0.798        & 0.954       & 0.953   & 0.953        & 0.973        & 0.969    & 0.969         \\ \hline
\end{tabular}
\label{tab:performance_metrics}
\end{table*}

\subsubsection{Controllability:} As shown in Table~\ref{tab:performance_metrics}, \arch demonstrated strong alignment between intended and generated content across all dimensions. Knowledge extraction precision and recall were 0.791 and 0.787 respectively, indicating \arch accurately generates messages according to specified knowledge without missing important content. Method alignment showed consistent performance across topics, with EDA videos performing slightly better. Action and Interaction components showed progressively stronger performance, demonstrating \arch's ability to reliably convert specified interactions into conversational utterances.

The ascending performance pattern (Knowledge → Method → Action → Interaction) reflects the natural refinement process as the system moves from general knowledge to specific interactions, leveraging work done in previous phases.
\vspace{-2em}
\subsubsection{Component Importance:} The full condition has the highest TrueSkill score ($\mu = 30.14, \sigma = 1.60$), suggesting that combining both knowledge induction and method planning is the most effective strategy. Knowledge-Only ($\mu = 26.11, \sigma = 1.40$) outperformed Method-Only ($\mu = 23.89, \sigma = 1.40$). This suggests that without knowledge induction, the method-only approach may fail to provide precise guidance tailored to the video content, potentially leading to over-scaffolding simple concepts or under-scaffolding more complex ones. In contrast, while the knowledge-only condition has a higher score than the method-only approach, it lacks the adaptive capability provided by method planning, making it less responsive to the student's mastery level. The baseline condition performed worst ($\mu = 19.86, \sigma = 1.60$).

Kruskal-Wallis test indicates significant differences between conditions ($p < 0.001; H = 32.86$), and Dunn's post-hoc test confirmed significant pairwise differences between all ablations ($p < 0.05$). These results demonstrate that each component in \arch contributes significantly to generating effective and coherent guidance, and the full pipeline has the best results than other conditions.


%% file: 05_discussion_new.tex
\section{Discussion and Conclusion}

Our research demonstrates \arch's effectiveness in enhancing programming education through interactive video-based tutoring. The evaluation results confirm that \arch successfully integrates learning science principles from the \framework framework to generate targeted pedagogical interventions. The architecture's ability to segment videos by learning goals, maintain controllability across knowledge and pedagogical dimensions, and adapt instruction based on student performance addresses key challenges in LLM-based tutoring.

\arch shows particular promise in formal classroom settings, where instructors can leverage its controllable parameters to create customized learning tasks within video-based lessons. The architecture's strengths lie in its structured prompting approach that maintains pedagogical coherence while allowing for personalized guidance. By integrating BKT-based student modeling with generative AI, \arch represents a significant step toward scalable, adaptive programming education.

However, our current implementation has limitations that future work should address. While \arch performs well with the three video topics in our evaluation, its effectiveness across more diverse programming tutorials requires further investigation. The architecture's current design assumes videos have a modular structure where segments align with discrete learning goals—an assumption that may not hold for tutorials with overlapping content. Future versions could implement adaptive segmentation using temporal coherence and multimodal cues to better align with instructional goals.


%% file: 00_main.bbl
\begin{thebibliography}{10}
\providecommand{\url}[1]{\texttt{#1}}
\providecommand{\urlprefix}{URL }
\providecommand{\doi}[1]{https://doi.org/#1}

\bibitem{alario2016interactive}
Alario-Hoyos, C., Kloos, C.D., Est{\'e}vez-Ayres, I., Fern{\'a}ndez-Panadero, C., Blasco, J., Pastrana, S., Villena-Rom{\'a}n, J.: Interactive activities: the key to learning programming with moocs. Proceedings of the European Stakeholder Summit on Experiences and Best Practices in and Around MOOCs, EMOOCS  \textbf{319} (2016)

\bibitem{anderson2001taxonomy}
Anderson, L.W., Krathwohl, D.R.: A taxonomy for learning, teaching, and assessing: A revision of Bloom's taxonomy of educational objectives: complete edition. Addison Wesley Longman, Inc. (2001)

\bibitem{badrinath2021pybkt}
Badrinath, A., Wang, F., Pardos, Z.: pybkt: An accessible python library of bayesian knowledge tracing models. International Educational Data Mining Society  (2021)

\bibitem{Brown1989SituatedCA}
Brown, J.S., Collins, A.M., Duguid, P.: Situated cognition and the culture of learning. Educational Researcher  \textbf{18},  32 -- 42 (1989), \url{https://doi.org/10.3102/0013189X018001032}

\bibitem{cheng2024scientific}
Cheng, A.Y., Guo, M., Ran, M., Ranasaria, A., Sharma, A., Xie, A., Le, K.N., Vinaithirthan, B., Luan, S.T., Wright, D.T.H., Cuadra, A., Pea, R., Landay, J.A.: Scientific and fantastical: Creating immersive, culturally relevant learning experiences with augmented reality and large language models. In: Proceedings of the 2024 CHI Conference on Human Factors in Computing Systems. CHI '24, Association for Computing Machinery, New York, NY, USA (2024). \doi{10.1145/3613904.3642041}

\bibitem{collins1991cognitive}
Collins, A., Brown, J.S., Holum, A., et~al.: Cognitive apprenticeship: Making thinking visible. American educator  \textbf{15}(3),  6--11 (1991)

\bibitem{gagne2005principles}
Gagne, R.M., Wager, W.W., Golas, K.C., Keller, J.M., Russell, J.D.: Principles of instructional design (2005)

\bibitem{herbrich2007trueskill}
Herbrich, R., Minka, T., Graepel, T.: Trueskill(tm): A bayesian skill rating system. In: Advances in Neural Information Processing Systems 20. pp. 569--576. MIT Press (2007), \url{https://www.microsoft.com/en-us/research/publication/trueskilltm-a-bayesian-skill-rating-system/}

\bibitem{linn1995designing}
Linn, M.C.: Designing computer learning environments for engineering and computer science: The scaffolded knowledge integration framework. Journal of Science Education and technology  \textbf{4},  103--126 (1995)

\bibitem{liu2024teaching}
Liu, R., Zenke, C., Liu, C., Holmes, A., Thornton, P., Malan, D.J.: Teaching cs50 with ai: Leveraging generative artificial intelligence in computer science education. In: Proceedings of the 55th ACM Technical Symposium on Computer Science Education V. 1. p. 750–756. SIGCSE 2024, Association for Computing Machinery, New York, NY, USA (2024). \doi{10.1145/3626252.3630938}

\bibitem{10.1007/978-3-031-64302-6_19}
Ma, Q., Shen, H., Koedinger, K., Wu, S.T.: How to teach programming in the ai era? using llms as a teachable agent for debugging. In: Artificial Intelligence in Education: 25th International Conference, AIED 2024, Recife, Brazil, July 8–12, 2024, Proceedings, Part I. p. 265–279. Springer-Verlag, Berlin, Heidelberg (2024). \doi{10.1007/978-3-031-64302-6_19}

\bibitem{youdao_bcembedding_2023}
NetEase~Youdao, I.: Bcembedding: Bilingual and crosslingual embedding for rag. \url{https://github.com/netease-youdao/BCEmbedding} (2023)

\bibitem{10.1145/3544548.3581574}
Pardos, Z.A., Tang, M., Anastasopoulos, I., Sheel, S.K., Zhang, E.: Oatutor: An open-source adaptive tutoring system and curated content library for learning sciences research. In: Proceedings of the 2023 CHI Conference on Human Factors in Computing Systems. CHI '23, Association for Computing Machinery, New York, NY, USA (2023). \doi{10.1145/3544548.3581574}

\bibitem{Pea2004TheSA}
Pea, R.D.: The social and technological dimensions of scaffolding and related theoretical concepts for learning, education, and human activity. Journal of the Learning Sciences  \textbf{13}(3),  423--451 (2004). \doi{10.1207/s15327809jls1303_6}

\bibitem{ren2024human}
Ren, C., Pardos, Z.A., Li, Z.: Human-ai collaboration increases skill tagging speed but degrades accuracy. CoRR  \textbf{abs/2403.02259} (2024)

\bibitem{sheng-etal-2023-dialogue}
Sheng, Z., Finzel, R., Lucke, M., Dufresne, S., Gini, M., Pakhomov, S.: A dialogue system for assessing activities of daily living: Improving consistency with grounded knowledge. In: Muresan, S., Chen, V., Casey, K., David, V., Nina, D., Koji, I., Erik, E., Stefan, U. (eds.) Proceedings of the Third DialDoc Workshop on Document-grounded Dialogue and Conversational Question Answering. pp. 68--79. Association for Computational Linguistics, Toronto, Canada (2023). \doi{10.18653/v1/2023.dialdoc-1.8}

\bibitem{sonkar-etal-2023-class}
Sonkar, S., Liu, N., Mallick, D., Baraniuk, R.: {CLASS}: A design framework for building intelligent tutoring systems based on learning science principles. In: Bouamor, H., Pino, J., Bali, K. (eds.) Findings of the Association for Computational Linguistics: EMNLP 2023. pp. 1941--1961. Association for Computational Linguistics, Singapore (2023)

\bibitem{sreedhar-etal-2024-canttalkaboutthis}
Sreedhar, M.N., Rebedea, T., Ghosh, S., Zeng, J., Parisien, C.: {C}ant{T}alk{A}bout{T}his: Aligning language models to stay on topic in dialogues. In: Al-Onaizan, Y., Bansal, M., Chen, Y.N. (eds.) Findings of the Association for Computational Linguistics: EMNLP 2024. pp. 12232--12252. Association for Computational Linguistics, Miami, Florida, USA (2024). \doi{10.18653/v1/2024.findings-emnlp.713}, \url{https://aclanthology.org/2024.findings-emnlp.713}

\bibitem{sweller2021instructional}
Sweller, J.: Instructional design. In: Encyclopedia of evolutionary psychological science, pp. 4159--4163. Springer (2021)

\bibitem{upton2014dictionary}
Upton, G., Cook, I.: A dictionary of statistics 3e. Oxford University Press, USA (2014)

\bibitem{zhao2023survey}
Zhao, W.X., Zhou, K., Li, J., Tang, T., Wang, X., Hou, Y., Min, Y., Zhang, B., Zhang, J., Dong, Z., Du, Y., Yang, C., Chen, Y., Chen, Z., Jiang, J., Ren, R., Li, Y., Tang, X., Liu, Z., Liu, P., Nie, J.Y., Wen, J.R.: A survey of large language models (2024), \url{https://arxiv.org/abs/2303.18223}

\end{thebibliography}
